\documentclass[5p,times]{elsarticle}

\usepackage{float}
\usepackage{amssymb}
\usepackage[figuresright]{rotating}
\usepackage{svg}
\usepackage{hyperref}
\usepackage{comment}
\usepackage{booktabs}
\usepackage{amsmath}
\usepackage{cleveref}
\usepackage{caption}
\usepackage{color, colortbl}
\usepackage{dblfloatfix}
\usepackage{dirtree}
\newtheorem{definition}{Definition}

\begin{document}

\begin{frontmatter}

\title{Metric Privacy in Federated Learning for Medical Imaging: Improving Convergence and Preventing Client Inference Attacks}

\author[IFCA]{Judith Sáinz-Pardo Díaz\corref{correspondingauthor}}
\ead{sainzpardo@ifca.unican.es}
\author[INRIA]{Andreas Athanasiou}
\ead{andreas.athanasiou@inria.fr}
\author[INRIA]{Kangsoo Jung}
\ead{gangsoo.zeong@inria.fr}
\author[INRIA]{Catuscia Palamidessi}
\ead{catuscia@lix.polytechnique.fr}
\author[IFCA]{Álvaro López García}
\ead{aloga@ifca.unican.es}

\address[IFCA]{Instituto de Física de Cantabria (IFCA), CSIC-UC \\ Avda. los Castros s/n. 39005 - Santander (Spain)}
\address[INRIA]{INRIA Saclay, Ecole Polytechnique (LIX) \\ 1 Rue Honore d’Estienne d’Orves 91120 - Palaiseau, Ile-de-France (France)}
\cortext[correspondingauthor]{Corresponding author}

\begin{abstract}
Federated learning is a distributed learning technique that allows training a global model with the participation of different data owners without the need to share raw data. This architecture is orchestrated by a central server that aggregates the local models from the clients. This server may be trusted, but not all nodes in the network. Then, differential privacy (DP) can be used to privatize the global model by adding noise. However, this may affect convergence across the rounds of the federated architecture, depending also on the aggregation strategy employed. In this work, we aim to introduce the notion of metric-privacy to mitigate the impact of classical server side global-DP on the convergence of the  aggregated model. Metric-privacy is a relaxation of DP, suitable for domains provided with a notion of distance. We apply it from the server side by computing a distance for the difference between the local models.
We compare our approach with standard DP by analyzing the impact on six classical aggregation strategies. The proposed methodology is applied to an example of medical imaging and different scenarios are simulated across homogeneous and non-i.i.d clients.
Finally, we introduce a novel \emph{client inference attack}, where a semi-honest client tries to find whether another client participated in the training and study how it can be mitigated using DP and metric-privacy. 
Our evaluation shows that metric-privacy can increase the performance of the model compared to standard DP, while offering similar protection against client inference attacks. 

\end{abstract}


\end{frontmatter}

\section{Introduction}\label{sec:intro}

The development of artificial intelligence models, such as machine and deep learning (ML/DL) and their application on distributed data is a field of emerging interest, especially with the rise of \emph{federated learning} \cite{li2020review}. This architecture allows to train a global ML/DL model on multiple data owners without sharing their data with each other or with a third party \cite{mcmahan2017communication}. 

Federated learning (FL) is considered a privacy preserving machine learning architecture, since the global model is built from local models trained by each data owner (in the following, client), without the need for the data to leave the devices, institutions or data centers that generate and/or store it. However, these approaches are not free from attacks on the privacy of the clients involved and the data used, with the most notable being  membership inference attacks \cite{mia_attack, sia_paper}, reconstruction attacks and model inversion attacks \cite{DLG, chen2022practical} among others. 

Differential privacy (DP) and homomorphic encryption (HE)
are commonly included in FL architectures to add an additional layer of privacy. For example, if we want to perform a secure model aggregation, HE can be applied so that the models are aggregated in an encrypted way \cite{ma2022privacy}. Another option is to add noise to each local model before sending it for aggregation (applying DP), in case the aggregator (in the following, the server) is not trusted.

A variant of standard DP is \emph{metric-privacy} (also known as \textit{d-privacy}) \cite{dprivacy, alvim2018metric} which can be employed in domains with a notion of distance. Like standard DP, metric-privacy offers a bound on the probability that the same result is obtained from two different datasets. This probability depends on a parameter $\epsilon$ (like standard DP) but also on the distance between the two datasets (unlike standard DP). In other words, metric-privacy ensures that the noise is calibrated to achieve privacy within a specific ``radius'' of distance. This is especially beneficial in applications where hiding an element within a group of neighbors provides adequate privacy protection. However, metric-privacy has received little attention in FL. 

Protecting against an untrusted central server using DP has been widely studied in FL. However most studies focus only on the server using \emph{FedAvg} (Federated Average) as the aggregation function. Therefore, there is a gap in the literature regarding the impact of global-DP in the accuracy of the model when other aggregation functions are used. 

Moreover, in this paper we also pose a different problem together with its potential solution. Specifically,  we assume that all clients are semi-honest and that the server (i.e. aggregator) is trusted. However, we suppose that one of the clients can act as an attacker to find out if another one is participating in the training or not. We name this a \emph{Client Inference Attack} (CIA). Knowing that a specific client participated in the dataset can lead to further attacks, which can be a privacy concern. For instance, it may strengthen other attacks, such as the membership inference attack.

Our approach, which will be detailed in Section~\ref{sec:methodology}, is to use metric-privacy from the server side (global-metric-privacy) to protect the aggregated model from this type of attack. Hereafter we will refer to global-metric privacy just as metric privacy for simplicity of terminology. Note that using metric-privacy we can better adjust the added noise obtaining a better performance and convergence of the model (compared to standard DP), while simultaneously protecting against CIAs.  

For the experimental part of this work, an openly available dataset belonging to the field of medical imaging has been used, simulating different scenarios in terms of client distribution (homogeneous or non-i.i.d.), as well as an additional scenario to test the protection against the proposed client inference attacks.

\subsection{Contributions and structure of the work}
The main contributions of this work can be summarized as follows:
\begin{itemize}
    \item First, we apply global-DP (i.e. server-side DP) to test the performance of the model across six different aggregation strategies. 
    This yields interesting results as global-DP affects the performance of the model differently depending on the chosen aggregation function. We experimentally evaluate the performance of metric-privacy, global-DP and, as a baseline, vanilla FL (i.e. without adding any noise at all), across different aggregation functions, showing that metric-privacy offers in all cases a better model accuracy than global-DP. The experiments are conducted using a medical imaging database openly available and simulating different clients and scenarios. To the best of our knowledge this comparison between different aggregation strategies and privacy approaches has not been studied before in scientific papers. 
    
    \item We study the application of metric-privacy from a trusted server by introducing a proper distance metric. 
    This metric is dynamically calculated from the server side in each round only taking into account the model parameters (i.e. without analyzing the clients' data distributions, which could affect privacy). 
    As far as we are aware of, this is the first work that incorporates metric-privacy into an FL architecture. 
    
    \item We introduce the concept of \emph{Client Inference Attack} where a semi-honest client tries to  infer whether or not a certain client is participating in the training. We conduct experiments showing that while both metric-privacy and global-DP provide adequate protection against this type of attack, the former produces a more accurate model. 
\end{itemize}

This work is structured as follows: first, in Section~\ref{sec:sota} we report the state of the art and the related work. Section~\ref{sec:fl_aggregation} presents different aggregation strategies that can be used in federated learning. Then, Section~\ref{sec:dp_fl} provides the theoretical basis for using DP in federated learning. In Section~\ref{sec:ciattacks} we define the client inference attacks. Section~\ref{sec:methodology} explains the methodology followed and how we can apply metric-privacy in FL. Next, in Section~\ref{sec:experiments} we conduct experiments across homogeneous and non-i.i.d clients and a CIA scenario.
Finally, Section~\ref{sec:conclusions} draws the conclusions and outlines future work.

\section{Related work}\label{sec:sota}

The incorporation of DP in data science ecosystems, especially in relation to the training of ML/DL models, is a topic that has been extensively studied in the field of privacy-preserving machine learning \cite{pan2024differential}. Regarding the training of deep learning models (more specifically deep neural networks), the use of DP during the model training has been explored in multiple works following the differentially private stochastic gradient descent (DP-SGD) approach \cite{abadi2016deep, ha2019differential, xie2021differential}, which consists of adding noise during the process of stochastic gradient descent. This is implemented in different Python libraries, like TensorFlow Privacy \cite{tensorflow2015-whitepaper} or Opacus \cite{Opacus}. 

FL is susceptible to privacy attacks, despite the fact that each client trains their model locally, without revealing their local dataset.
Among various inference attacks, the Membership Inference Attack (MIA)  ~\cite{nasr2019comprehensive,mia_attack,carlini2022membership} determines whether a particular record is part of a dataset or not.  To do so, it is assumed that the adversary (central server) owns a so-called shadow model of the target client, which is similar to the target client’s model (i.e. follows the same distribution) but is trained on a different, disjoint, dataset.
MIAs are widely employed in the literature for evaluating privacy risks \cite{10.1145/3523273}.

The Source Inference Attack (SIA)~\cite{sia_paper} is a more advanced inference attack. An honest-but-curious server in FL attempts to identify exactly which client owns a specific data point used for training. To estimate the source of the data point, the server leverages the prediction losses of local models on particular data points by using a Bayesian approach.

Among reconstruction attacks, arguably the most well-known is Deep Leakage from Gradients (DLG) ~\cite{DLG}. DLG usually exploits the shared gradients during the training process in federated learning to reconstruct sensitive training data. This attack shows that even gradient information can leak important details about the original data. In both reconstruction and inference attacks, applying DP is crucial to protect sensitive information of the participating clients.

However, in FL adversaries may also take the form of participants or other external actors with access to the (shared) aggregated model. In the case of an FL architecture where the server that orchestrates the training is not trusted, clients can add DP during training by applying DP-SGD, but they can also do so once the model is trained, adding noise to the model updates \cite{10386466}. This will prevent the central server from extracting information from the data used in the training. 

Concerning medical imaging, in \cite{adnan2022federated} the authors apply DP-SGD in a FL architecture for medical image analysis, specifically using histopathology images. On the other hand, several works have explored the incorporation of DP to FL architectures. For example, in \cite{9082603} DP is added from each client to the model updates before sending the model to the server, specifically in a personalized federated learning approach.

In this work we assume that the server is trusted. The attacker can either be another client who tries to infer information about other clients (e.g. find out if a certain client is participating in the training). Hence the model that we aim to protect is the aggregated (global) one. Therefore, the server can globally add DP to the aggregated model after receiving the updates from all the clients. This will prevent the extraction of information by the clients in each round. Also it will protect the final resulting model when it is made publicly available.
Few works can be found in the literature concerning applying DP from the server side. For example, in \cite{naseri2020local} local and central DP are compared in a FL setting showing the protection against backdoor attacks, defining central DP (CDP) as the case in which \textit{``the FL aggregation function is perturbed by the server''}. In this work we refer to this approach as \emph{global-DP}.

\section{Federated learning aggregation functions}\label{sec:fl_aggregation}

The classic horizontal federated learning architecture involves a server and multiple clients or data owners. On the one hand, the clients have their data stored locally in their own infrastructures and are in charge of performing the model training in a distributed way. On the other hand, the server is in charge of aggregating the individual locally trained models in order to build a global model. Thus, the scheme followed under a FL architecture can be summarized in the following steps: (1) each client receives a model to be trained locally (defined by the server or consensuated by the clients); (2) each client trains the model locally with its data; (3) the clients send the parameters or weights that define the model trained locally (or the model updates); (4) the server receives the models/updates from the clients involved and aggregates them to build a global model; (5) the aggregated global model is distributed back to the clients to be retrained. This process is repeated from step (2) for a pre-defined number of rounds. 

One of the most critical steps is the aggregation process carried out from the server side, since it is the moment when the global model is built from all the locally trained models.
Thus, selecting an appropriate aggregation strategy in relation to the use case employed is key to achieve an aggregated global model that is robust and accurate. Some classically applied aggregation functions are presented below:

\paragraph{Federated Average (FedAvg)}

The most natural approach is to simply average the parameters. However, it must be taken into account that there will be clients who have trained the model using more data than others, thus being able to obtain a more accurate prediction. Thus, it is important that the average is made in a weighted way taking into account the number of data available for each client. Proposed by B. McMahan et al. in 2017 (\cite{mcmahan2017communication}) when introducing the FL architecture (where its successful application was shown for the MNIST, CIFAR10 and Shakespeare datasets), the Federated Averaging strategy (\textit{FedAvg}) is the most widely used aggregation function in the literature for its simplicity and effectiveness. 

Let $n$ be the number of clients  participating in the FL architecture, $n_i$ the number of data of client $i$ ($\forall i \in \{1,\hdots,n\}$) and $w_i^{(r)}$ the weights or model parameters obtained for each client $i$ after training in the round $r$. The aggregation performed with \textit{FedAvg} consists on the weighted mean of the parameters of each clients, as follows, for round $r$:

\begin{equation}\label{eq:fedavg}
    w^{(r)} = \frac{1}{\sum_{i=1}^{n}n_{i}}\sum_{i=1}^{n}w_{i}^{(r)}n_{i}
\end{equation}

Then, $w^{(r)}$ are the aggregated weights resulting from round $r$ that will be transmitted to each client when starting round $r+1$. 

\paragraph{Federated Average with Momentum (FedAvgM)}

This strategy was implemented in \cite{fedavgm} and the proposed \textit{FedAvgM} method arises from the idea of improving stability and efficiency by using the momentum to store gradients from previous rounds, also in order to dampen oscillation in the training. Then, let $w^{(r)}$ the global parameters in round $r$ and $\Delta w^{(r+1)}$ the aggregated difference between the parameters obtained in rounds $r+1$ and $r$: $\Delta w^{(r+1)}=\frac{1}{\sum{i=1}^{n}n_{i}}\sum_{i=1}^{n} n_{i}(w^{(r)}-w_{i}^{(r+1)})$, note that $w_{i}^{(r+1)}$ are the local parameters obtained for client $i$ in round $r$.

In order to get the aggregated parameters using \textit{FedAvgM} we have to calculate the momentum vector in round $r$ $v^{(r)}$ (initialized as $v^{(0)}=0$), in such a way that $v^{(r+1)}=\beta v^{(r)} + \Delta w^{(r+1)}$. Then, we will calculate the weights in round $r+1$ as $w^{(r+1)}=w^{(r)}-v^{(r+1)}$, being $\beta$ the momentum. 
Additionally, we can add the term $\mu$ as the server learning rate introducing it as follows: $w^{(r+1)}=w^{(r)}-\mu v^{(r+1)}$.

\textit{FedAvgM} aims to incorporate the idea of momentum at the server side during the aggregation process to improve the stability and speed of convergence of the overall model \cite{fedavgm}. According to the implementation in the \textit{Flower} library \cite{beutel2020flower}, if the value for the momentum ($\beta$) is strictly greater than zero, some global initial parameters for the model must be included. As will be explained in Section~\ref{sec:experiments}, a validation dataset will be used to get these initial model parameters.

\paragraph{Federated Median (FedMedian)}

The Federated Median strategy (\textit{FedMedian}) was introduced in \cite{fedmedian}. The proposed algorithm for aggregating using the median takes into account byzantine machines so that only in the case of normal worker machines each such client computes the local gradient. Thus, let $g_{i}(w^{(r)})$ be the gradients obtained for client $i$ with the weights aggregated in round $r$ ($w_{r}$). The central server will be in charge of getting the median of the local gradients in each round $r$: $g_{median}(w^{(r)})=median(g_{i}(w^{(r)}), i\in\{1,\hdots,n\})$. Then, once such aggregated gradient is calculated, the weights are updated for round $r+1$: $w^{(r+1)}=w^{(r)}-\mu g_{median}(w^{(r)})$ \cite{fedmedian}, being $\mu$ the step-size or learning rate.      

\paragraph{FedProx}

The main objective of the \textit{FedProx} aggregation strategy is to deal with heterogeneous settings in FL schemes \cite{fedprox}. Specifically, it is a generalization and re-parametrization of the \textit{FedAvg} strategy. In order to calculate $w_{i}^{(r+1)}$ (the weights for client $i$ in round $r+1$) we have to minimize the following objective function $h_{i}$ for each client $i$, with $F_{i}(\cdot)$ the global objective function at each client $i$:

$$
h(w, w^{(r)})=F_{i}(w) + \frac{\mu}{2}||w-w^{(r)}||^{2}.
$$

Thus, $w_{i}^{(r+1)}=arg \min_{w} h_{i}(w)$. Then the server aggregates the values $w_{i}^{(r+1)}$ $\forall i \in \{1,\hdots,n\}$ by taking the mean. 

Note than in this approach we need to choose a value for the parameter $\mu$.

\paragraph{Federated Optimization (FedOpt) and adaptive Federated Optimization with Yogi (FedYogi)}

The \textit{FedOpt} strategy introduced in Algorithm 1 from \cite{reddi2020adaptive} aims to improve \textit{FedAvg} by using gradient-based optimizers with given learning rates that must be customized both from the client and server side. Specifically, let $w^{(r)}$ the global weights at round $r$ and $w_{i}^{(r)}$ the weights for client $i$ at round $r$, then we can define $\Delta w^{(r)}_{i} = w_{i}^{(r)} - w^{(r)}$ and $\Delta w^{(r)} = \frac{1}{n}\sum_{i=1}^{n}\Delta w^{(r)}_{i}$. From the server side, we will use an optimizer for calculating $w^{(r+1)}$ given $w^{(r)}$, $\Delta w^{(r)}$ and the learning rate. 

Note that this approach allows the use of different adaptive optimizers in the server side, such us \textit{Adam}, \textit{Adagrad} or \textit{Yogi} among others (the we will have the \textit{FedAdam}, \textit{FedAdagrad} and \textit{FedYogi} aggregation strategies respectively). The adaptive methods are used on the server side while SGD is used from the client side \cite{reddi2020adaptive}. More details are given in the pseudocode of Algorithm 2 from \cite{reddi2020adaptive}.

These strategies require additional parameters, such as the server-side and client-side learning rate, $\beta_{1}$ and $\beta_{2}$ as momentum and second momentum parameters respectively and $\tau$ for controlling the degree of adaptability of the algorithm. According to its implementation in Flower, some initial global parameters for the model need to be introduced when using these functions. The use of these global initial parameters provides a consistent starting point for all clients.

\section{Differentially private federated learning}\label{sec:dp_fl}

Differential privacy (DP) is a privacy preserving technique that aims to provide a formal guarantee about what an analyst (adversary) can learn about an individual in a database. The probability that the adversary observes  any event is comparable in cases where certain information from an individual is or is not included in the dataset.  In addition, with DP we can ensure that an adversary with unlimited computational capacity and auxiliary information cannot break the established privacy level.

In other words, DP states that an algorithm is differentially private if by viewing its result an adversary cannot know whether a particular individual's data is included in the database or not. 
This is typically done with the addition of noise (i.e. data obfuscation).
The goal is to protect user's privacy while allowing the data to be meaningful for analysis. For this purpose, different mechanisms can be followed to ensure that the added noise does not significantly alter the analysis.

Let's start by introducing the notions of adjacency and randomized algorithm which are crucial for defining DP:

\begin{definition}
    Two databases $\mathcal{D}$ and $\overline{D}$ are adjacent (notated by $\mathcal{D} \sim \mathcal{D'}$) if they differ by exactly one record.
\end{definition}

\begin{definition}
    A probabilistic or randomized algorithm for the query $f$ over the database $\mathcal{X}$ is a probabilistic function $g$ from $\mathcal{X}$ to a set of values $\mathcal{Z}$ such as $g:\mathcal{X} \longrightarrow \mathcal{D(Z)}$, with $\mathcal{D(Z)}$ the set of probability distributions in $\mathcal{Z}$.    
\end{definition}

Now we can define $\epsilon$\textit{-differential privacy}:

\begin{definition}
A randomized algorithm $\mathcal{M}$, with domain $\mathcal{D}$ and range $\mathcal{R}$, satisfies $\epsilon$-differential privacy if for every pair of adjacent datasets $D,\overline{D} \in \mathcal{D}$, for every $S \subseteq \mathcal{R}$ and $\epsilon>0$ :

$$
\mathbb{P}[\mathcal{M}(D)\in S] \leq e^{\epsilon}\mathbb{P}[\mathcal{M}(\overline{D})\in S]
$$    
\end{definition}

In view of the above definition, the value of $\epsilon$ is the \emph{privacy budget}, which allows to control the level of privacy (the amount of privacy loss allowed). We  note that the lower the value of $\epsilon$, the higher the privacy, but, in most of the cases, the lower the usefulness of the data for analysis (although this is not always the case \cite{natasha_dp_e_utility}).

 Approximate DP is a famous relaxation of the previous definition where we allow the above inequality to not hold with a small probability. We can define $(\epsilon,\delta)$\textit{-differential privacy} as:

\begin{definition}
A randomized algorithm $\mathcal{M}$, with domain $\mathcal{D}$ and range $\mathcal{R}$, verifies $(\epsilon,\delta)$-differential privacy if for every two pair of adjacent datasets $D,\overline{D} \in \mathcal{D}$ and for every $S \subseteq \mathcal{R}$, $\epsilon>0$ and $\delta\in [0,1]$:

$$
\mathbb{P}[\mathcal{M}(D)\in S] \leq e^{\epsilon}\mathbb{P}[\mathcal{M}(\overline{D})\in S] +\delta
$$    
\end{definition}
 
The parameter $\delta$ is the probability of exceeding the privacy budget, i.e. with probability $1-\delta$ the privacy loss will not be greater than $\epsilon$.

Some of the most commonly used mechanisms to ensure DP are  among others the Laplace Mechanism, the Exponential Mechanism and the Gaussian Mechanism. 

Let's introduce the Gaussian Mechanism, as it will be used during the experimental settings performed in this work.

First, let us define the notion of sensitivity:

\begin{definition}
Be $f:\mathcal{D} \longrightarrow \mathbb{R}^{k}$, the $l_{2}$-sensitivity of $f$ is defined as follows:

$$
\displaystyle
\Delta_{2}(f):= \max_{||x-y||_1}||f(x)-f(y)||_{2}
$$
\end{definition}

Now we can define the Gaussian Mechanism as given in Definition~\ref{def:gaussian}:
\begin{definition}\label{def:gaussian}
Given the function $f:\mathcal{D} \longrightarrow \mathbb{R}^{k}$, we define the Gaussian Mechanism ($\mathcal{M}_{G}$) for ($\epsilon$, $\delta$)-differential privacy as follows:
$$
\displaystyle
\mathcal{M}_{G}(x, f(\cdot), \epsilon, \delta):= f(x) + (Y_{1}, \hdots, Y_{k}).
$$
Note that $Y_{i}$ ($\forall i \in \{1, \hdots, k\}$) follows the distribution $N(0, \sigma^2)$, with $\sigma = \frac{\Delta_{2}(f) \sqrt{2\log(1.25/\delta)}}{\epsilon}$.
\end{definition}

DP can be applied using two approaches: locally and globally. In the local setting we assume that the clients have to obfuscate their data on their own, using  local-DP (LDP). On the other hand, in the global setting the clients totally trust a central entity to collect their data and add noise to the output of a query or algorithm, using global-DP (GDP). Note that trusting the curator or central server is a vital assumption. If such assumption cannot be made using local-DP is a better choice. 

If we extrapolate these notions to a FL architecture, we can think of two paradigms depending on the side of the network in which the potential attacker is located:

\begin{itemize}
    \item \textit{Trusted server:} in this case we can add DP from the server side. The goal here is to protect against some clients that can act as attackers, or to prevent information extraction from the final global model when it is published. This is exactly our aim in this work.
    
    In this line, the idea is to perturb the global aggregated model by adding noise to it once aggregated. In this work, we analyze the impact of adding such noise according to the aggregation strategy applied, assuming that we have a network composed by a trusted honest server. 

    \item \textit{Untrusted server:} in this case the clients will be interested in sending to the server a noisy version of the model. This is because some information could be extracted concerning the data of each client by analyzing the local updates that are sent in each round. In this line the clients can train the model and then add noise before sending to the server for aggregating, or to perform differentially private stochastic gradient descent (DP-SGD) for adding DP during the training process. Recall that we do not consider this scenario in this work.
\end{itemize}

Another variation of DP is  \textit{metric-privacy} or \textit{d-privacy} \cite{dprivacy, galli2023group}. In metric-privacy the datasets are not just adjacent (i.e.  they just have one value that is not the same) but their actual difference is calculated using some distance metric. Then, the privacy level can be adapted so as it offer better privacy when the distance is small. In other words, metric-privacy allows the adversary to find a rough estimation of the clients' data, but the actual values remain protected. This is done in exchange of better utility, as less noise is added.

We can define metric-privacy as:

\begin{definition}

Let $\mathcal{D}$ be a domain provided with a metric $d:\mathcal{D}^{2}\longrightarrow \mathbb{R}_{\geq 0}$. 
The randomized algorithm $\mathcal{M}$ with range $\mathcal{R}$ and with domain $\mathcal{D}$ satisfies $\epsilon$-metric-privacy, if for every two pair of inputs $x,\overline{x} \in \mathcal{D}$ with distance $d(x,\overline{x})$, every $y \in \mathcal{R}$ and every $\epsilon>0$:

$$
\mathbb{P}[\mathcal{M}(x)=y] \leq e^{\epsilon\cdot d(x,\overline{x})}\mathbb{P}[\mathcal{M}(\overline{x})=y]
$$    
\end{definition}

In this work our goal is to evaluate the impact of metric-privacy and classic global-DP according to different aggregation strategies in FL architectures. Note that our goal is to guarantee metric-privacy separately for each round of the FL architecture. 

Finally, note that we aim to analyze in practice how can we improve the convergence of a federated model using global metric-privacy (in the following metric-privacy or metric-DP) with respect to classic global-DP, while we are protecting the aggregated model from information extraction and ensuring a similar protection regarding client inference attacks. We want to highlight that this work doesn't aim to perform a fair comparison between global-DP and metric-privacy (as the amount of noise added in each approach is different, as will be explained later), but to introduce this notion for better tuning the amount of noise added for finding a balance between predictive performance and resistance against client inference attacks.

\section{Client Inference Attacks}\label{sec:ciattacks}

One of the main hypotheses that we want to assess within this work is that metric-privacy, besides offering a better accuracy than standard global-DP in FL, may also be helpful against inference attacks.


In this section, we introduce the
\emph{Client Inference Attack (CIA)}, a novel attack where a semi-honest client (attacker) receives the global aggregated model from an honest (trusted) server and tries to determine if another client belongs to the list of participants. This attack can serve as a first step to strengthen other attacks, such as a Membership Inference Attack \cite{mia_attack} or a Source Inference Attack \cite{sia_paper}. 


For instance, consider a model jointly built by hospitals and competing pharmaceutical companies to treat a disease. No party (apart from the central server) knows exactly which hospitals or which companies are working on the model. The central server is honest and releases the aggregated model back to the clients. However, a curious  pharmaceutical company $P$ may want to launch a CIA attack and figure out whether or not a certain hospital $H$ is participating in the training. If this CIA shows that $H$ did indeed participated, then $P$ can facilitate further inference attacks on $H$ and the data used (which may be associated with the patients from the hospital).

In order to achieve this attack, we assume that the attacker has sufficient knowledge of the target client’s training dataset to create a \emph{shadow training dataset}; a dataset that mimics the distribution of the target client’s dataset. This shadow dataset can either be entirely disjoint from the target dataset (but sampled from the same distribution) or it may overlap as a subset, if the attacker has obtained some data points from the target.
Note that a similar assumption is essential to other well-studied inference attacks such as the Membership Inference Attack \cite{mia_attack}. In addition, the clients participating in a FL architecture can also receive information about the process from the server, such as the aggregated metrics calculated in each round, in order to evaluate the convergence of the model.

Let us consider a bit vector $s$ with $n$ bits, with each bit $i$ being set to 1 if user $i$ participated in some round $r$ in the training of the model $W$, and 0 otherwise. We can define the CIA as follows:

\begin{definition}[Client Inference Attack] Given a global aggregated model $W$ and the shadow dataset $D_x$ of client $x$, the Client Inference Attack can be defined as:

$$
C(D_{x},W):= \mathbb{P}(s_{x} =1 | D_{x}, W)
$$
\end{definition}

Note that in this work we assume that the number of clients is fixed in all rounds. This makes sense in the case of cross-silo configurations. Thus, the client that is going to perform a CIA, wants to know if the target has participated or not based on the shadow dataset available.

First of all, it is clear that global-DP can be a measure that helps to prevent this type of attack. This is because in CIA the malicious client analyzes the performance of the aggregated model. Adding Differential Privacy to the aggregated model limits the amount of information disclosed by mathematical definition. Thus, adding noise to the aggregated model seeks to mask the individual contributions of the different participating clients. However, it is necessary to calibrate the noise added to maintain a good balance between protection against CIA and the performance of the model. 

In this sense, we propose to use the notion of metric-privacy. As we are about to explore, adding noise based on a metric allows to make a finer balance between privacy and model performance, since in case of similar contributions from the clients it will be necessary to add less noise. Even in scenarios with non-i.i.d. clients, metric-privacy can contribute to the proper calibration of privacy by adding only the ``necessary amount'' of noise. Using metric-privacy in this case makes sense as we assume that the participating clients are static throughout the rounds.

We will show in \Cref{sec:experiments} an experimental analysis of this approach in an example of medical imaging. The hypothesis which we will be testing is that metric-privacy can contribute to better calibrate the noise compared to global-DP. Such noise calibration can reduce the defense against CIA, but at the same time improve the performance of the same allowing to achieve a better balance between utility and privacy.


\section{Proposed methodology}\label{sec:methodology}

We are now ready to describe our proposed approach. Let us recall our trust model:
we assume that there is an honest central server  in a cross silo setting with all the initial clients participating during the whole training. However, we suppose that some of the clients involved may act as attackers seeking to extract information from the models and therefore from the data of other clients participating in the network. Specifically, our goal is to protect against Client Inference Attacks.

To do this, we first apply differential privacy with fixed clipping norm to the aggregation process from the server side (global-DP). By doing so we also prevent information leakage from external attackers, when publishing the aggregated model in a model repository. 
Specifically, in our approach we use the Gaussian Mechanism setting a noise multiplier associated to the privacy level we want to define.

Concerning the selection of the parameters for applying DP, we have to note that the clipping norm is a hyperparameter that must be optimized. More specifically, 
for the first round we calculate the following metric:

$$
\tilde{c} = \max_{\substack{i,j\in \{1,\hdots,n\} \\ i\neq j}}||w_{i}-w_{j}||_{2}.
$$

We started optimizing the clipping norm using $\tilde{c}$, and we finally take a value $C=5$ for the experimental simulations after performing different tests.

Next, we propose a novel adaptation of metric-privacy to deal against such attacks. 
Recall that with metric-privacy we aim to offer privacy guarantees based on the distance between the clients.
However, the server does not have access to the raw data of the clients. Hence the question arises: \emph{how can the server calculate the distance between the clients, without knowing their data?}

To overcome the hurdle, we propose a metric that depends on the distance between the model updates rather than the actual clients' data.
In order to do so, the server calculates the maximum distance for each pair of clients by analyzing the local weights received from each one. 

Let $w_{i}^{(n)}$ be the local weights of the model trained for client $i$ in round $n$, with $w_{i}^{(n)}(\ell)$ being the weights of the layer $\ell$ of the model in round $n$. Moreover let $\mathcal{L}$ be the set of layers of the trained model and $n_{c}$ be the number of clients. We define  the distance in round $n$  as follows:

$$
d^{(n)} = \max_{\substack{i,j\in \{1,\hdots,n_{c}\} \\ i\neq j}} \left(\frac{1}{|\mathcal{L}|}\sum_{\ell \in \mathcal{L}} ||w_{i}^{(n)}(\ell)-w_{j}^{(n)}(\ell)||_{F} \right).
$$

Note that $||\cdot||_{F}$ represents the Frobenius norm. 

The hypothesis which we will be testing is that using metric privacy with $d^{(n)}$ will help to fine-tune the amount of noise needed in each round, ultimately allowing better convergence. 

We explained above that we will compare metric privacy with global-DP applied using the Gaussian mechanism. For a fair comparison we will employ the Gaussian mechanism in metric privacy as well setting $\overline{\epsilon}=\epsilon \cdot d^{(n)}$ for each round $n$.

The classic implementation of DP with fixed clipping in FL involves introducing the clipping norm $C$, the noise multiplier ($n_{\epsilon}$) and the number of sampled clients $n_{c}$. Note that the noise multiplier has the opposite interpretation to the privacy budget, since the lower the value, the higher the utility and the lower the privacy, as less noise is injected. 

In this line, we have calculated the metric $d^{(n)}$ dynamically in each round $n$ and use it by dividing the standard deviation of the Gaussian noise as follows:

$$
\mathcal{N}\left(0,\frac{n_{e}\cdot C}{n_{c} \cdot d^{(n)}}\right).
$$

Then, the objective is to analyze the impact of adding metric-privacy to the aggregated model form the server side, ensuring that metric-privacy in fulfilled in each FL round. In addition, we want to evaluate the impact of this privacy enhancing technology compared to classic global-DP in terms of a client inference attack. Note that if $d^{(n)}>1$, we are adding more noise with metric-privacy if we keep the same noise multiplier, but we want to study how this method allow us to find a proper value for the noise added while giving similar protection against client inference attacks. This will be studied experimentally in the following section.

\section{Experimental results}\label{sec:experiments}

\subsection{Data under study}\label{sec:data}

In this work, for the experimental simulation of the proposed methodology  we have taken the data from the Alzheimer dataset available in \cite{alzheimer_mri_dataset}. Our goal is to carry out this comparative study with this open dataset that can be meaningful in a practical scenario. The dataset contains brain MRI images and is especially useful for the testing and investigation of applied healthcare AI models. Specifically, the goal with this dataset is to predict, based on the MRI images, the level of Alzheimer's disease of the patient. For this purpose there are 4 categories: mild demented (0), moderate demented (1), non-demented (2) and very mild demented (3). The initial train dataset is composed of 5120 images and in the test dataset 1280 images are available. The distribution of the classes in each of these two datasets is shown in Table~\ref{tab:original_data}. 

\begin{table}[ht]
    \centering
    \begin{tabular}{cccccc}
    \toprule
         & \textit{\textbf{Total}} & \textit{\textbf{Class 0}} & \textit{\textbf{Class 1}} & \textit{\textbf{Class 2}} & \textit{\textbf{Class 3}} \\
         \midrule
         \textit{Train} & 5120 & 724 & 49 & 2566 & 1781 \\
         \textit{Test} & 1280 & 172 & 15 & 634 & 459 \\
    \bottomrule
    \end{tabular}
    \caption{Original train and test set distribution.}
    \label{tab:original_data}
\end{table}

\begin{figure}[ht]
    \centering
    \includegraphics[width=\linewidth]{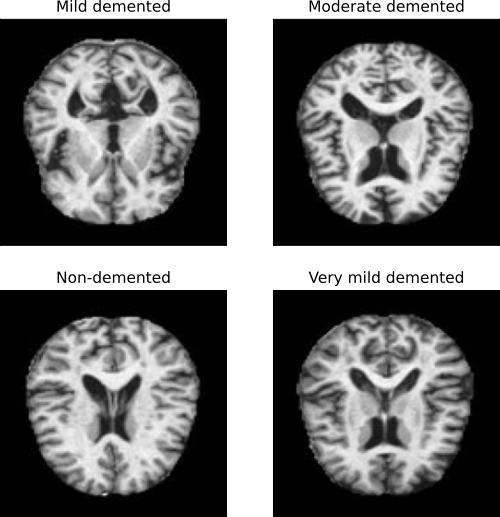}
    \caption{Example of the original images from the train set by classes.}
    \label{fig:example_images}
\end{figure}

In addition, Figure~\ref{fig:example_images} shows an example of each class of images available in the train dataset.

To perform the FL scheme we will simulate 4 clients from the train dataset, while leaving the initial test set for analyzing the performance of the final models and for validation. Specifically, the 50$\%$ of the test set will be taken as validation from the server side to train some initial parameters for those aggregation functions that require them. The remaining 50$\%$ will be considered the test set for analyzing the resulting models.

In order to analyze the impact of adding global-DP and metric privacy according to the aggregation function used in FL, we propose two experimental settings for the same initial centralized dataset: homogeneous clients and non-i.i.d clients. Once the train data has been distributed to the different clients, in both cases we leave 20$\%$ for test in each client and 80$\%$ for train. This split is done in a stratified way for maintaining the statistical significance associated with the representation of each class.

\subsubsection{Homogeneous clients}\label{sec:clients_iid}

In this first approach we perform a division of the initial training set into four disjunct clients such that the distribution of the number of images in each class is homogeneous. For this, we perform a stratified division of the training set by assigning to each client the same number of data in a balanced way (almost the same number of each of each class for each client split). Thus, Table~\ref{tab:homogeneous_distribution} shows for each client the distribution of the four classes present in the original database.

\begin{table}[ht]
    \centering
    \begin{tabular}{cccccc}
    \toprule
         & \textit{\textbf{Total}} & \textit{\textbf{Class 0}} & \textit{\textbf{Class 1}} & \textit{\textbf{Class 2}} & \textit{\textbf{Class 3}} \\
         \midrule
         \textit{Client 1} & 1280 & 181 & 12 & 641 & 446 \\
         \textit{Client 2} & 1280 & 181 & 12 & 642 & 445 \\
         \textit{Client 3} & 1280 & 181 & 12 & 642 & 445 \\
         \textit{Client 4} & 1280 & 181 & 13 & 641 & 445 \\
    \bottomrule
    \end{tabular}
    \caption{Homogeneous clients distribution.}
    \label{tab:homogeneous_distribution}
\end{table}

\subsubsection{Non-i.i.d clients}\label{sec:clients_noniid}

In this second setting, the aim is to simulate a scenario where we work with clients who possess different datasets. For this purpose, the initial training set has been divided among four disjoint clients, with each assigned a different number of data points and an unequal distribution of classes represented in the dataset. Specifically, Table~\ref{tab:noniid_distribution} shows the number of data for each client and the number of data belonging to each predicting class in each of them.

\begin{table}[ht]
    \centering
    \begin{tabular}{cccccc}
    \toprule
         & \textit{\textbf{Total}} & \textit{\textbf{Class 0}} & \textit{\textbf{Class 1}} & \textit{\textbf{Class 2}} & \textit{\textbf{Class 3}} \\
         \midrule
         \textit{Client 1} & 1792 & 280 & 16 & 881 & 615\\
         \textit{Client 2} & 768 & 107 & 13 & 368 & 280\\
         \textit{Client 3} & 2048 & 257 & 17 & 1054 & 720\\
         \textit{Client 4} & 512 & 80 & 3 & 263 & 166\\
    \bottomrule
    \end{tabular}
    \caption{Non-i.i.d clients distribution.}
    \label{tab:noniid_distribution}
\end{table}

\subsection{Model analyzed}\label{sec:model}

Different models and hyperparameters have been analyzed during the experimental procedure detailed in this work in order to select the one implemented. Finally, the implemented model for performing the different tests (which was fine-tuned taking into account a validation split of the dataset) is a convolutional neural network (CNN) composed by the following layers:

\begin{itemize}
    \item \textit{Conv2D layer}. 32 neurons. Activation \textit{relu}. Input shape: \textit{(128, 128, 1)}.
    \item \textit{MaxPooling2D((2,2))}.
    \item \textit{Conv2D layer}. 64 neurons. Activation \textit{relu}.
    \item \textit{MaxPooling2D((2,2))}.
    \item \textit{Conv2D layer}. 128 neurons. Activation \textit{relu}.
    \item \textit{Flatten layer}.
    \item \textit{Dense layer}. 64 neurons. Activation \textit{relu}.
    \item \textit{Dropout layer}. Rate 0.1.
    \item \textit{Dense layer}. 32 neurons. Activation \textit{relu}.
    \item \textit{Dropout layer}. Rate 0.1.
    \item \textit{Output: Dense layer}. 4 neurons. Activation \textit{softmax}.
\end{itemize}

The model is compiled using the \textit{Adam} optimizer, the \textit{sparse categorical cross-entropy} as loss function and the \textit{accuracy} as metric. The batch size used for the training in 32 and we train it during 5 epochs and 20 rounds of the Federated Learning architecture.

Note that in order to follow a real use case, we have performed a hyperparameter tuning using a validation dataset. Then, we have run the model to simulate a real use case with a fixed seed for the libraries with stochastic behavior, in order to make a fairer comparison between the three approaches (non-DP, global-DP and metric-privacy). However, in order to analyze the robustness of the results, we have tested the performance by running the model 5 times with each scenario and each aggregation function (e.g. for the case of i.i.d. clients we have conducted 90 experiments). These results are shown for the test set as a function of the accuracy in Table~\ref{tab:5runs_iid} in \ref{sec:5runs}. In view of the standard deviation obtained in the 5 runs of the model for such scenario, which is statistically significant, the results of one run of the model are shown to better fit a real scenario (see for instance \cite{adnan2022federated} or \cite{HATAMI2024128119}). 

\subsection{Results}\label{sec:results}

In this section we show the results obtained with the three approaches: vanilla FL (baseline, no DP), global-DP and metric-privacy from the server side with six different aggregation strategies. In addition, we show the aggregated accuracy obtained for the clients test set and accuracy, precision and F1-score obtained with each strategy for the initial test set. 
Finally, we compare the three approaches  regarding the prevention of the Client Inference Attack.

\subsubsection{Homogeneous clients}\label{sec:results_iid}

The aggregated results in the last five rounds of the FL architecture (mean and standard deviation), using the CNN model presented in the previous section are shown in Table~\ref{tab:cnn_results}, comparing the mean accuracy and standard deviation in the last 5 rounds across the six aggregation strategies. 

\begin{table}[ht]
    \centering
    \resizebox{\linewidth}{!}{
    \begin{tabular}{cccc}
         \toprule
          & \multicolumn{3}{c}{\textbf{Mean accuracy $\pm$ std (last 5 rounds)}} \\
         \cmidrule{2-4}
         \textbf{Strategy} & \textit{Vanilla FL} & \textit{Global-DP} & \textit{Metric-privacy}\\
         \midrule
         \textit{FedAvg} & 0.929 $\pm$ 0.008 & 0.881 $\pm$ 0.013 & 0.905 $\pm$ 0.007 \\ 
         \textit{FedAvgM} & 0.878 $\pm$ 0.013 & 0.759 $\pm$ 0.024 & 0.799 $\pm$ 0.022 \\ 
         \textit{FedMedian} & 0.920 $\pm$ 0.005 & 0.885 $\pm$ 0.013 & 0.905 $\pm$ 0.018  \\ 
         \textit{FedProx} & 0.931 $\pm$ 0.006 & 0.879 $\pm$ 0.008 & 0.907 $\pm$ 0.022 \\ 
         \textit{FedOpt} & 0.956 $\pm$ 0.004 & 0.918 $\pm$ 0.014 & 0.948 $\pm$ 0.005 \\ 
         \textit{FedYogi} & 0.950 $\pm$ 0.006 & 0.906 $\pm$ 0.008 & 0.911 $\pm$ 0.012 \\ 
         \bottomrule
    \end{tabular}}
    \caption{Comparison on homogeneous clients of the mean and the standard deviation of the aggregated accuracy in the last five rounds of the FL scheme. Aggregation performed without using DP and with global-DP and metric-privacy with noise multiplier of 0.01. }
    \label{tab:cnn_results}
\end{table}

\begin{table*}[!t]
    \centering
    \resizebox{\linewidth}{!}{
    \begin{tabular}{c ccc c ccc c ccc}
         \toprule
         & \multicolumn{3}{c}{\textit{Vanilla FL}} & & \multicolumn{3}{c}{\textit{Global-DP}} & & \multicolumn{3}{c}{\textit{Metric-privacy}}\\
         \cmidrule{2-4} \cmidrule{6-8} \cmidrule{10-12}
         \textbf{Strategy} & \textit{\textbf{Accuracy}} & \textit{\textbf{F1-Score}} & \textit{\textbf{Precision}} &  &  \textit{\textbf{Accuracy}} & \textit{\textbf{F1-Score}} & \textit{\textbf{Precision}} &  & \textit{\textbf{Accuracy}} & \textit{\textbf{F1-Score}} & \textit{\textbf{Precision}}\\
         \midrule
         \textit{FedAvg} & 0.909 & 0.903 & 0.950 &  & 0.884 & 0.862 & 0.918 &  & 0.894 & 0.843 & 0.943 \\ 
         \textit{FedAvgM} & 0.884 & 0.768 & 0.932 &  & 0.762 & 0.554 & 0.607 &  & 0.823 & 0.608 & 0.631 \\ 
         \textit{FedMedian} & 0.932 & 0.899 & 0.946 &  & 0.875 & 0.827 & 0.913 &  & 0.895 & 0.843 & 0.943 \\ 
         \textit{FedProx} & 0.909 & 0.848 & 0.873 &  & 0.877 & 0.853 & 0.914 &  & 0.903 & 0.877 & 0.944 \\ 
         \textit{FedOpt} & 0.950 & 0.936 & 0.970 &  & 0.918 & 0.877 & 0.911 &  & 0.930 & 0.917 & 0.950 \\ 
         \textit{FedYogi} & 0.933 & 0.914 & 0.957 &  & 0.908 & 0.841 & 0.884 &  & 0.917 & 0.888 & 0.941 \\ 
         \bottomrule
    \end{tabular}}
    \caption{Accuracy, F1-score and precision obtained in the test set with each approach. Homogeneous clients.}
    \label{tab:accuracy_test}
\end{table*}
Table~\ref{tab:cnn_results} shows that with metric-privacy we always achieve better results compared to those with global-DP, as well as a better convergence over the course of the rounds (see \ref{sec:agg_accuracy}). The evolution of the accuracy of the aggregated model in clients test set (aggregated) is shown for each round of the FL schema in Figure~\ref{fig:accuracy_iid} from \ref{sec:agg_accuracy} for the cases of the global aggregated model without DP, with global-DP and with metric-privacy.

Both in the case of global-DP and metric-privacy we introduce the same value for the noise multiplier (0.01), however, note that with metric-privacy we are dividing the noise multiplier by the distance $d^{(n)}$ in each round $n$. If $d^{(n)}>1$ we will be adding more noise (we could expect the model to get worse in its performance), while if $d^{(n)}<1$ we would be adding less noise, which will result in greater similarity with the non-DP model. Then, metric-privacy help us to find a proper value for the standard deviation of the Gaussian noise, while we are still protecting privacy and client inference attacks, as will be shown in Section~\ref{sec:results_ciattacks}.

In Table~\ref{tab:accuracy_test} we show the results obtained with each approach and each aggregation function for the global model obtained after 20 rounds for the test set in terms of the accuracy, F1-score and precision. In view of Table~\ref{tab:accuracy_test} we can highlight that the performance with metric-privacy is better than the one with global-DP with all the six strategies analyzed in terms of the accuracy and the precision, and in 5 out of 6 in terms of the F1-score. In addition, note that for \textit{FedProx} strategy, the F1-score and the precision is better with metric-privacy that without DP, this may be due to a better generalization capability in terms of the less-represented classes and a better mitigation of potential overfitting.

In addition, for completeness of the analysis, \ref{sec:plots_auc} shows the value for the AUC and the ROC curve obtained in each case with each strategy and each of the three proposed approaches (Figure~\ref{fig:auc_iid}).

\subsubsection{Non-i.i.d clients}\label{sec:results_noniid}

Here, we conduct the experiments on the non-i.i.d. clients case. 
It should be noted that the distribution of each class in each data owner is statistically significant with the initial distribution, as it would not make sense for a hospital to have the minority class as the majority class, which would not fit the usual distribution of the pathology.

The non-i.i.d. nature is determined by the amount of data in each data owner, with one being more representative than the others as a function of the number of samples, as shown in Table~\ref{tab:noniid_distribution}. Then, we have trained the FL architecture with these four clients and the six aggregation strategies explained above. Note that some of these strategies will be more appropriate than others to deal with the non-i.i.d nature of the simulated data owners. 

Table~\ref{tab:cnn_results_noniid} shows the aggregated results obtained for each client's test set in terms of the mean accuracy in the last 5 rounds of the FL scheme as well as the standard deviation in these rounds.

\begin{table}[ht]
    \centering
    \resizebox{\linewidth}{!}{
    \begin{tabular}{cccc}
         \toprule
          & \multicolumn{3}{c}{\textbf{Mean accuracy $\pm$ std (last 5 rounds)}} \\
        \cmidrule{2-4}
         \textbf{Strategy} & \textit{Vanilla FL} & \textit{Global-DP} & \textit{Metric-privacy}\\
         \midrule
         \textit{FedAvg} & 0.933 $\pm$ 0.002 & 0.878 $\pm$ 0.012 & 0.927 $\pm$ 0.008 \\ 
         \textit{FedAvgM} & 0.909 $\pm$ 0.007 & 0.790 $\pm$ 0.018 & 0.840 $\pm$ 0.016 \\ 
         \textit{FedMedian} & 0.926 $\pm$ 0.009 & 0.838 $\pm$ 0.013 & 0.905 $\pm$ 0.007  \\ 
         \textit{FedProx} & 0.937 $\pm$ 0.014 & 0.878 $\pm$ 0.016 & 0.943 $\pm$ 0.003 \\ 
         \textit{FedOpt} & 0.964 $\pm$ 0.007 &  0.925 $\pm$ 0.009 & 0.945 $\pm$ 0.006 \\ 
         \textit{FedYogi} & 0.945 $\pm$ 0.005 & 0.912 $\pm$ 0.009 & 0.919 $\pm$ 0.010 \\ 
         \bottomrule
    \end{tabular}}
    \caption{Comparison on non-i.i.d clients of the mean and the standard deviation of the aggregated accuracy in the last five rounds of the FL scheme. Aggregation performed without using DP and with global-DP and metric-privacy with noise multiplier of 0.01. }
    \label{tab:cnn_results_noniid}
\end{table}

From Table~\ref{tab:cnn_results_noniid} we can see that the results with metric-privacy are again better than those obtained with classic global-DP in the 6 cases analyzed depending on the aggregation function. Moreover, as we have already seen in the case of homogeneous clients, we can observe that in the case of \textit{FedProx} strategy, the mean accuracy with metric-privacy is even better than without adding DP. Although again this may be due to a reduction of overfitting for better adjustment of the underrepresented class, we can highlight by observing the standard deviation, that in the best case without DP we would have an accuracy of 0.951, while in the best case with metric-privacy it would be 0.946. Therefore, this improvement is due to fluctuations in the overall aggregated accuracy in the last 5 rounds, as can be seen in Figure~\ref{fig:accuracy_noniid} of \ref{sec:agg_accuracy}. 

\begin{table*}[t]
    \centering
    \resizebox{\linewidth}{!}{
    \begin{tabular}{c ccc c ccc c ccc}
         \toprule
         & \multicolumn{3}{c}{\textit{Vanilla FL}} & & \multicolumn{3}{c}{\textit{Global-DP}} & & \multicolumn{3}{c}{\textit{Metric-privacy}}\\
         \cmidrule{2-4} \cmidrule{6-8} \cmidrule{10-12}
         \textbf{Strategy} & \textit{\textbf{Accuracy}} & \textit{\textbf{F1-Score}} & \textit{\textbf{Precision}} &  &  \textit{\textbf{Accuracy}} & \textit{\textbf{F1-Score}} & \textit{\textbf{Precision}} &  & \textit{\textbf{Accuracy}} & \textit{\textbf{F1-Score}} & \textit{\textbf{Precision}}\\
         \midrule
         \textit{FedAvg} & 0.919 & 0.930 & 0.956 &  & 0.867 & 0.787 & 0.886 &  & 0.925 & 0.914 & 0.931 \\ 
         \textit{FedAvgM} & 0.909 & 0.862 & 0.944 &  & 0.784 & 0.573 & 0.594 &  & 0.837 & 0.621 & 0.640 \\ 
         \textit{FedMedian} & 0.933 & 0.917 & 0.961 &  & 0.859 & 0.626 & 0.641 &  & 0.917 & 0.926 & 0.940 \\ 
         \textit{FedProx} & 0.920 & 0.910 & 0.950 &  & 0.886 & 0.871 & 0.904 &  & 0.903 & 0.890 & 0.909 \\ 
         \textit{FedOpt} & 0.936 & 0.927 & 0.957 &  & 0.922 & 0.913 & 0.960 &  & 0.916 & 0.885 & 0.889 \\ 
         \textit{FedYogi} & 0.927 & 0.870 & 0.950 &  & 0.900 & 0.845 & 0.836 &  & 0.895 & 0.847 & 0.865 \\ 
         \bottomrule
    \end{tabular}}
    \caption{Accuracy, F1-score and precision obtained in the test set with each approach. Non-i.i.d clients.}
    \label{tab:accuracy_test_noniid}
\end{table*}

In addition, Table~\ref{tab:accuracy_test_noniid} shows the results obtained with the six aggregation strategies and the three paradigms (no DP or vanilla FL, global-DP and metric-privacy) on the test dataset. Specifically, this table shows the model performance in this test set as a function of the accuracy, F1-score and precision obtained with the aggregated model obtained after 20 rounds of FL training. In this case, again the trend is maintained: metric-privacy improves the global-DP results, in general. Metric privacy is better in all cases with respect to accuracy and precision and in 5 out of 6 with respect to F1-score (only slightly worse with \textit{FedAvg}). Furthermore, with \textit{FedProx} we see the anomaly already commented (i.e. the precision and F1-score obtained with metric-privacy is better than without DP).

Finally, for further analysis, Figure~\ref{fig:accuracy_noniid} in \ref{sec:agg_accuracy} shows the evolution of the aggregated accuracy in the training set of the different clients in this scenario of non-i.i.d. clients with the three proposed privacy approaches. In addition Figure~\ref{fig:auc_noniid} in \ref{sec:plots_auc} shows the value for the AUC and the ROC curve obtained in each case for the test set.

\subsection{Impact of a client inference attack}\label{sec:results_ciattacks}

In the last two sections we have seen that by including metric-privacy in the FL architecture, from the server side, we can improve the accuracy of the model, as we are better tuning the amount of noise added. However, we are interested in analyzing whether adding metric-privacy can contribute to reduce the risk of a client inference attack (following the definition proposed in Section~\ref{sec:ciattacks}) in a similar way as with global-DP. 

Specifically, as already stated, in this case our assumption is that a client can act as an attacker to infer whether certain client is participating in the training, and thus learn information about it. For example, in this medical imaging case, an attacker might want to infer whether a certain hospital is participating in the training, in order to learn about the hospital's data distribution and thus sensitive information about the hospital's patients and their pathologies. 

In this sense, we can intuitively expect that the attack will be more effective in the first round. Although an attacker would use additional information to try to extract additional insights throughout the different rounds, it will be in the first round (when the model has not yet converged) when it will be possible to infer more knowledge. As a reminder, we assume that the clients participating in the federated training are the same throughout all rounds.

Thus, we are going to simulate a new scenario. In this case, for the sake of simplicity, we will take 3 clients, one of which will be the attacker who wants to know whether or not another client is participating in the training. In addition, this client on which the attack will be performed  (target) will have an anomalous data distribution compared to the distribution on the other two clients.

In Table~\ref{tab:distribution_ciattack} the distribution of the clients simulated for this experiment is shown, with client 1 being the attacker and client 3 being the target. Note that in this case, in order to get a better fit to the training data of each client, we are going to train the model for 20 epochs. Then, at the end of the first round, client 1 (attacker) knows its local model as well as the global aggregated one and some aggregated metrics sent from the server. 

\begin{table}[ht]
    \centering
    \begin{tabular}{cccccc}
    \toprule
         & \textit{\textbf{Total}} & \textit{\textbf{Class 0}} & \textit{\textbf{Class 1}} & \textit{\textbf{Class 2}} & \textit{\textbf{Class 3}} \\
         \midrule
         \textit{Client 1} & 1747 & 120 & 9 & 1122 & 496 \\ 
         \textit{Client 2} & 1591 & 180 & 11 & 894 & 506 \\ 
         \textit{Client 3} & 1882 & 524 & 29 & 550 & 779 \\ 
    \bottomrule
    \end{tabular}
    \caption{Clients distribution for simulating a client inference attack.}
    \label{tab:distribution_ciattack}
\end{table}

For our approach concerning CIA from client 1 to client 3 as target, we assume that the former has a shadow dataset from client 3, which will be a random split of 10$\%$ of client 3 training data. In addition, as the attacker participates in the training he/she will receive from the server the aggregated metrics (accuracy and loss) obtained in each round. Then, we can analyze the loss obtained with the aggregated model for the shadow train dataset of client 3. In Table~\ref{tab:ciattacks_example} we show the results for the cases where this client participates in the training performing the aggregation with the \textit{FedAvg} approach (as it is the most used one) and the three privacy approaches.

\begin{table}[ht]
    \centering
    \begin{tabular}{cccc}
    \toprule
         \textit{\textbf{Client}} & \textit{\textbf{Vanilla FL}} & \textit{\textbf{Global-DP}} & \textit{\textbf{Metric-privacy}} \\
         \midrule
         \textit{Aggregated} & 1.032 & 3.603 & 1.135 \\ 
         \textit{Target}  & 1.182 & 4.848 & 1.506 \\ 
         \midrule
         \textit{Difference ($\%$)} & 12.719  & 25.679  & 24.631 \\
    \bottomrule
    \end{tabular}
    \caption{Loss obtained for the aggregated test set and for the shadow dataset of client 3 with \textit{FedAvg} strategy.}
    \label{tab:ciattacks_example}
\end{table}

Note that in Table~\ref{tab:ciattacks_example} we take into account the loss instead of other metrics such as the accuracy, as for this multi-class classification problem it is more representative, as we are dealing with clients with quite non-i.i.d classes distributions. In view of this example, we can note that for the \textit{FedAvg} strategy we get quite similar values for the relative difference between the aggregated result in each client's test sets and the loss for the target client shadow dataset with global-DP and with metric-privacy. As we could expect, we get a slightly greater relative difference with global-DP than with metric-privacy, but in both cases we are protecting from the CIA attacks in a similar way (more than in the case with no DP (vanilla FL), with a 12.719$\%$ of difference), with a difference of less than 1.1$\%$ between global-DP and metric-privacy approaches. 

In Table~\ref{tab:ciattacks_example_fedopt} we also show the results obtained for this CIA simulation using the \textit{FedOpt} strategy, as this is the one that provides better results in general for the previous scenarios analyzed. We can observe that in this case the convergence of the model is much better due to the aggregation strategy used, since it has an auxiliary initial model trained with the validation dataset. Also, in this case it is worth noting that we obtain a greater difference with metric-privacy than with global-DP, thus providing a greater resistance to the CIA in this particular example. At the same time, as is shown in Table~\ref{tab:cnn_results_cia}, the results for the loss of the test set are also better with metric-privacy than with global-DP. 

\begin{table}[ht]
    \centering
    \begin{tabular}{cccc}
    \toprule
         \textit{\textbf{Client}} & \textit{\textbf{Vanilla FL}} & \textit{\textbf{Global-DP}} & \textit{\textbf{Metric-privacy}} \\
         \midrule
         \textit{Aggregated} & 0.595 & 0.743 & 0.756 \\ 
         \textit{Target}  & 0.599 & 0.807 & 0.853 \\ 
         \midrule
         \textit{Difference ($\%$)} & 0.598 & 7.936 & 11.320 \\ 
    \bottomrule
    \end{tabular}
    \caption{Loss obtained for the aggregated test set and for the shadow dataset of client 3 with \textit{FedYogi} strategy.}
    \label{tab:ciattacks_example_fedopt}
\end{table}

Then, as analyzed in the previous sections, in this use case we reach a better balance between protection and model performance with metric-privacy, as in all cases analyzed in Sections~\ref{sec:results_iid} and \ref{sec:results_noniid} we obtain a significantly better model performance than with global-DP. In addition, in view of the above we get a similar protection from CIA attacks with \textit{FedAvg} (slightly worse than with global-DP), and better for the case of \textit{FedOpt} for this new scenario. For completeness of the analysis, \ref{sec:cia_results} shows the results obtained for the other four aggregation functions analyzed.

Finally, let's see the results for the loss obtained in the test set with these three clients trained after the first round (see Table~\ref{tab:cnn_results_cia}):

\begin{table}[ht]
    \centering
    \begin{tabular}{cccc}
         \toprule
          & \multicolumn{3}{c}{\textbf{Test loss}} \\
         \cmidrule{2-4}
         \textbf{Strategy} & \textit{Vanilla FL} & \textit{Global-DP} & \textit{Metric-privacy}\\
         \midrule
         \textit{FedAvg} & 1.010 & 3.514 & 1.106 \\ 
         \textit{FedAvgM} & 0.993 & 1.048 & 1.041 \\ 
         \textit{FedMedian} & 1.046 & 2.071 & 1.387 \\ 
         \textit{FedProx} & 1.010 & 1.744 &  1.149  \\ 
         \textit{FedOpt} &  0.645  & 0.845  & 0.823 \\ 
         \textit{FedYogi} & 0.870 & 1.036 & 0.902 \\ 
        \bottomrule
    \end{tabular}
    \caption{Comparison of the loss obtained for the test set in the first round. Aggregation performed without using DP, with global-DP with noise multiplier of 0.01 and with metric-privacy with 0.01 as noise multiplier. CIA scenario.}
    \label{tab:cnn_results_cia}
\end{table}

In view of Table~\ref{tab:cnn_results_cia}, we can see that in all cases the loss obtained with metric-privacy is lower than with global-DP, while the protection against CIA is similar, and in some cases even higher (e.g. with \textit{FedProx} and \textit{FedAvgM}, see \ref{sec:cia_results}). We have to note that the model is not yet converging, as only one round of the scheme has been carried out. However, it is important to analyze the results after this round because it is the one from which an attacker could extract more information. Again in this simple approach the best results are obtained with \textit{FedOpt} as aggregation strategy.

\section{Conclusions and future work}\label{sec:conclusions}

In this paper we have analyzed the impact of including the notion of metric-privacy in an FL architecture. Specifically, we have focused on the case where a trusted server and a fixed set of semi-honest clients are available and can act as attackers in the network to extract information. 

Additionally, we present a new type of attack, the client inference attack (CIA), in which a client in the network aims to infer whether or not a certain client is participating in the architecture. This information can serve as a basis for further attacks, such as reconstruction, source or membership attacks.

In this sense, we propose the use of metric-privacy versus the use of classic global-DP, so that the noise added in each round is calibrated according to a distance metric calculated as the maximum distance between the models of each pair of clients. Thus, we guarantee metric-privacy in each round, tuning the amount of noise added (allowing for better model performance), and at the same time offer certain guarantees against CIA. 

Specifically, we analyze a use case of medical imaging in three cases using 6 FL aggregation functions: homogeneous clients, non-i.i.d clients and a third case with a client with unbalanced data (target), in order to test the resistance to CIA. We have observed that in all cases the metric-privacy approach performance is significantly better than the global-DP approach, while offering a similar resistance to CIA in most cases, and even better in 2 out of 6 for the third case. 

This work has been applied to an openly available dataset of medical images, but future work includes its applicability to other types of use cases to analyze the impact of the proposed methodology, which aim to improve the performance of the FL architectures while protecting privacy. Note that this is the first work that we are aware of that employs metric-privacy in a federated architecture to improve privacy/utility trade-off. At the same time it is the first one that defines the CIA and presents an example of application and prevention with global-DP and metric-privacy. Future work includes the application to new case studies in different fields, as well as the analysis of different CIA scenarios and the study of their impact and usefulness as a basis for other attacks.\\

\noindent \textbf{CRediT authorship contribution statement}
\noindent \textbf{Judith Sáinz-Pardo Díaz}: Conceptualization, Formal analysis, Investigation,   Methodology, Software, Validation, Visualization, Writing - Original Draft. 
\textbf{Andreas Athanasiou}: Conceptualization, Investigation, Methodology,  Writing - Original Draft.
\textbf{Kangsoo Jung}: Conceptualization, Investigation, Methodology,  Writing - Original Draft.
\textbf{Catuscia Palamidessi}: Conceptualization, Methodology,  Writing - Original Draft, Supervision.
\textbf{Álvaro López García}: Conceptualization, Funding acquisition, Writing - Original Draft, Supervision.\\

\noindent \textbf{Declaration of Competing Interest}

\noindent The authors declare that they have no known competing financial interests or personal relationships that could have appeared to influence the work reported in this paper.\\

\noindent \textbf{Data availability}

\noindent The data used in this study are openly available and can be obtained from \cite{alzheimer_mri_dataset}.\\

\noindent \textbf{Acknowledgments}

\noindent Judith Sáinz-Pardo Díaz and Álvaro López García would like to thank the funding through the project AI4EOSC ``Artificial Intelligence for the European Open Science Cloud'' that has received funding from the European Union's Horizon Europe research and innovation programme under grant agreement number 101058593 and from the SIESTA project ``Secure Interactive Environments for Sensitive daTa Analytics'', funded by the European Union (Horizon Europe) under grant agreement number 101131957.

\noindent The work of Andreas Athanasiou is supported by the project CRYPTECS, funded by the ANR (project number ANR-20-CYAL-0006) and by the BMBF (project number 16KIS1439). The work of
Kangsoo Jung is supported by the project ELSA, funded by the Horizon Europe Framework (with project number 101070617). The work of Catuscia Palamidessi is supported by the project HYPATIA, funded by the ERC (grant agreement number 835294).

\bibliographystyle{elsarticle-num}
\bibliography{preprint}

\clearpage
\onecolumn
\appendix

\section{Analysis of the performance of the trained model in different runs}\label{sec:5runs}

In this section we show the mean accuracy obtained for the three approaches (vanilla FL, global-DP and metric-privacy) with the different aggregation strategies for the case of homogeneous clients. The results shown in Table~\ref{tab:5runs_iid} correspond with the mean and standard deviation of five runs of the models evaluated in the test dataset (90 models are evaluated in total, five runs for each strategy and privacy approach). Our idea is to evaluate the statistical significance of the trained model.

\begin{table}[ht]
    \centering
    \begin{tabular}{cccc}
         \toprule
          & \multicolumn{3}{c}{\textbf{Mean accuracy $\pm$ std (5 runs)}} \\
         \cmidrule{2-4}
         \textbf{Strategy} & \textit{Vanilla FL} & \textit{Global-DP} & \textit{Metric-privacy}\\
         \midrule
         \textit{FedAvg} & 0.930 $\pm$ 0.011 & 0.881 $\pm$ 0.013 & 0.913 $\pm$ 0.019 \\ 
         \textit{FedAvgM} & 0.880 $\pm$ 0.004 & 0.781 $\pm$ 0.021 & 0.812 $\pm$ 0.028 \\ 
         \textit{FedMedian} & 0.931 $\pm$ 0.000 & 0.867 $\pm$ 0.013 & 0.897 $\pm$ 0.004 \\ 
         \textit{FedProx} & 0.923 $\pm$ 0.011 & 0.880 $\pm$ 0.014 & 0.919 $\pm$ 0.011 \\ 
         \textit{FedOpt} & 0.947 $\pm$ 0.005 & 0.912 $\pm$ 0.010 & 0.924 $\pm$ 0.010 \\
         \textit{FedYogi} & 0.939 $\pm$ 0.008 & 0.899 $\pm$ 0.007 & 0.895 $\pm$ 0.013 \\ 
         \bottomrule
    \end{tabular}
    \caption{Mean and standard deviation of the accuracy obtained for the test set in 5 different runs of the model. Homogeneous clients,}
    \label{tab:5runs_iid}
\end{table}

Note that the interest of running the experiments multiple times is from the point of view of the two privacy approaches (that add a component of randomness, as well as other strategies that use an initial model). In all the cases a seed is fixed in order to start with the same model in all the clients (we see that with \textit{FedMedian} without DP the std is 0). In view of the standard deviation obtained in these 90 experiments (5 for each approach and aggregation function), for the other scenarios (non-i.i.d and client inference attacks), we show the results obtained for a fixed seed, for a better analysis of the results and for a better fit to a real use case where the model is run once the hyperparameter optimization has been completed and then it is evaluated on a new test dataset. We can note than in mean all the results with metric-privacy are better than with global-DP for the test set except for \textit{FedYogi}. However, if we take the upper bound taken into account the std, metric-privacy even performs slightly better than global-DP also in this case.

\section{Evolution of the aggregated accuracy in the clients' test set}\label{sec:agg_accuracy}

\begin{figure}[ht]
    \centering
    \includegraphics[width=\linewidth]{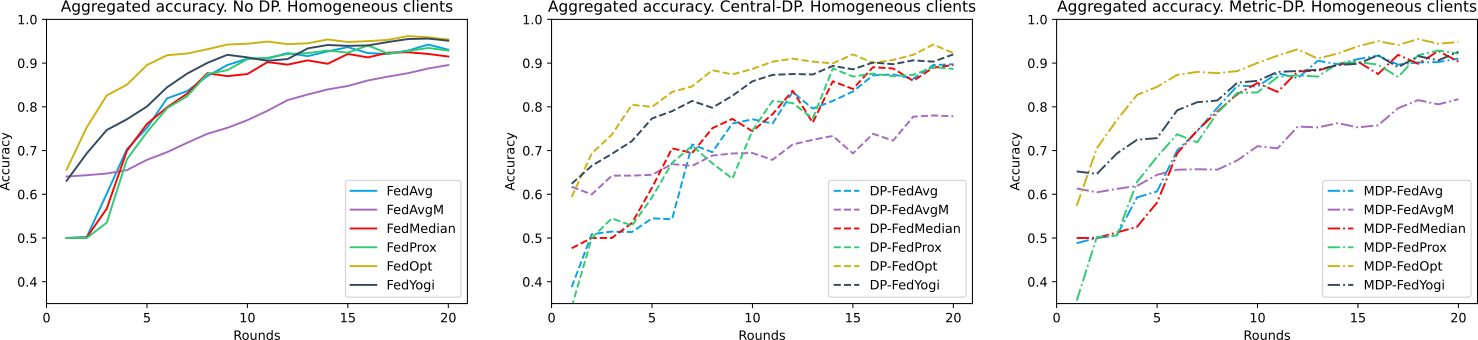}
    \caption{Evolution of the aggregated accuracy in each round of the FL architecture. Homogeneous clients.}
    \label{fig:accuracy_iid}
\end{figure}

In Figure~\ref{fig:accuracy_iid} the aggregated accuracy for each client's test set in each round of the FL scheme is shown for the case of homogeneous clients. Here we can note that the models training with metric-privacy from the server side reach a convergence more similar and softer to the model aggregated without DP that the one aggregated applying global-DP (which shows more fluctuations over the course of the rounds). It should be noted that in the three cases the trends of the rounds without DP, with global-DP and with metric-privacy (in the figure referred as MDP) are similar, with \textit{FedOpt} being the most accurate, followed by \textit{FedYogi} (which is more clearly observed with global-DP), and ending in all cases with \textit{FedMedian} and \textit{FedAvgM} as the least reliable.

\begin{figure}[ht]
    \centering
    \includegraphics[width=\linewidth]{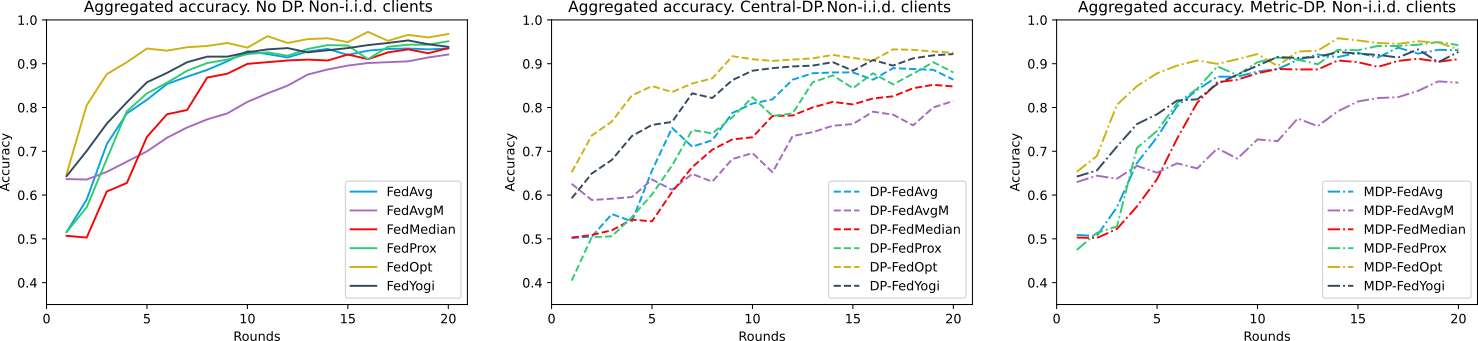}
    \caption{Evolution of the aggregated accuracy in each round of the FL architecture. Non-i.i.d clients.}
    \label{fig:accuracy_noniid}
\end{figure}

Then, Figure~\ref{fig:accuracy_noniid} shows the results obtained for the aggregated accuracy in each round for the case of non-i.i.d clients. Again in this case the curves are smoother with metric-privacy than with global-DP, due to the amount of noise added. However, it is interesting to note that with \textit{FedAvg}, in this scenario, better results are obtained in the first round, which may be counter-intuitive, but may be due to the fact that the aggregated results are shown and the client with more data (client 2) has more weight than the other clients in these aggregated results, as well as a greater weight in the aggregation of the model with this strategy. It is interesting to note that in all three cases the same trend is maintained, with \textit{FedOpt} being the best performing strategy, while \textit{FedAvgM} is the least accurate. 

\section{ROC curves and AUC in the test set}\label{sec:plots_auc}

\begin{figure}[H]
    \centering
    \includegraphics[width=\linewidth]{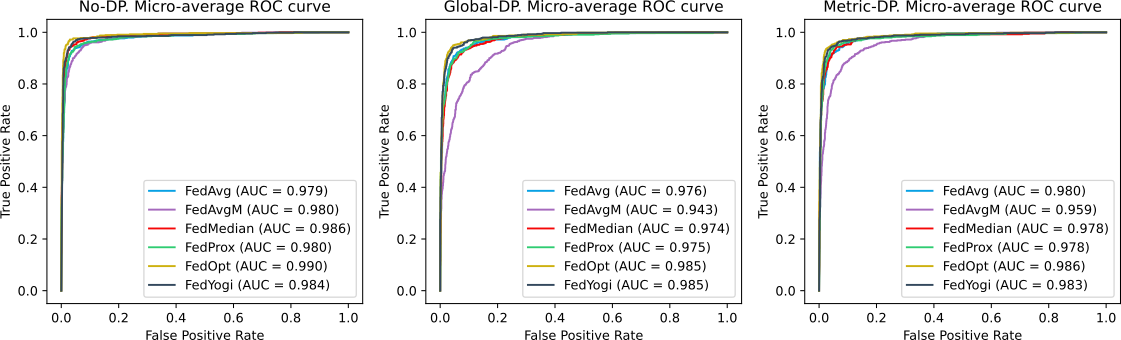}
    \caption{ROC curves and AUC obtained in the client test set with each strategy and DP approach. Homogeneous clients.}
    \label{fig:auc_iid}
\end{figure}

Figure~\ref{fig:auc_iid} shows the ROC curves and the AUC obtained with each aggregation strategy for the case of homogeneous clients and with the three approaches (no-DP or vanilla FL, global-DP, metric-privacy).

\begin{figure}[ht]
    \centering
    \includegraphics[width=\linewidth]{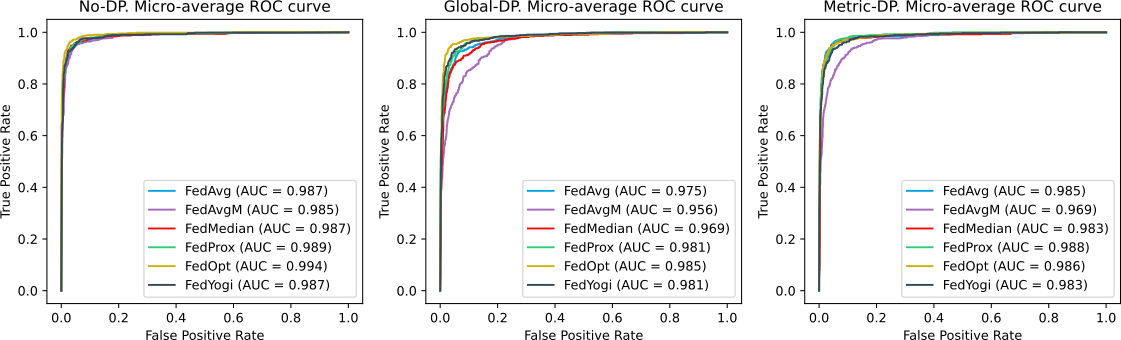}
    \caption{ROC curves and AUC obtained in the client test set with each strategy and DP approach. Non-i.i.d clients.}
    \label{fig:auc_noniid}
\end{figure}

In view of Figure~\ref{fig:auc_iid}, we can see that the best results for the scenario of homogeneous clients are reached without DP in 5 out of 6 cases (with \textit{FedAvg} the results are slightly better with metric-privacy, as explained in Section~\ref{sec:results_iid}). In addition, the ones obtained with metric-privacy are always better than those with global-DP. The former is also fulfilled for the case of non-i.i.d. clients as is shown in Figure~\ref{fig:auc_noniid}, but in this case the best results are always obtained without DP.

Note than in order to get the ROC curves and to calculate the AUC in this case of multi-class classification, the micro-averaged one-vs-rest ROC-AUC score has been used, comparing each class against the other three in the labels.

\section{Client inference attack for different aggregation strategies}\label{sec:cia_results}

In Section~\ref{sec:results_ciattacks} we have shown the performance obtained for the scenario simulated concerning the client inference attack performed using both \textit{FedAvg} and \textit{FedOpt} as aggregation strategies. For completeness of the analysis, the results for the other four aggregation functions are shown in this appendix.

We can note that fot the case of \textit{FedAvgM}, metric-privacy provides a greater difference that global-DP, while it also provides better results in terms of the loss. This may be due for a reduction in the overfitting to the target client. On the other hand, for the other three aggregation functions, a better result in terms of the distance is obtained with global-DP, but there are still better with metric-privacy that without DP. This shows that metric-privacy helps to prevent client inference attacks while allowing a better convergence of the model than classic global-DP.
\begin{table}[ht]
    \centering
    \begin{minipage}[t]{0.45\textwidth}
    \centering
    \begin{tabular}{cccc}
    \toprule
         \textit{\textbf{Client}} & \textit{\textbf{Vanilla FL}} & \textit{\textbf{Global-DP}} & \textit{\textbf{Metric-privacy}} \\
         \midrule
         \textit{Aggregated} & 0.897 & 0.969 & 0.945 \\ 
         \textit{Target}  & 1.118 & 1.115 & 1.211 \\ 
         \midrule
         \textit{Difference ($\%$)} & 19.782 & 13.105 & 21.965 \\ 
    \bottomrule
    \end{tabular}
    \caption{Loss obtained for the aggregated test set and for the shadow dataset of client 3 with \textit{FedAvgM} strategy.}
    \label{tab:ciattacks_example_fedavgm}
    \end{minipage}
    \hspace{0.05\textwidth} 
    \begin{minipage}[t]{0.45\textwidth}
    \centering
    \begin{tabular}{cccc}
    \toprule
         \textit{\textbf{Client}} & \textit{\textbf{Vanilla FL}} & \textit{\textbf{Global-DP}} & \textit{\textbf{Metric-privacy}} \\
         \midrule
         \textit{Aggregated} & 1.068 & 2.079 & 1.393 \\ 
         \textit{Target}  & 1.188 & 2.911 & 1.880 \\ 
         \midrule
         \textit{Difference ($\%$)} & 10.129 & 28.587 & 25.888 \\ 
    \bottomrule
    \end{tabular}
    \caption{Loss obtained for the aggregated test set and for the shadow dataset of client 3 with \textit{FedMedian} strategy.}
    \label{tab:ciattacks_example_fedmedian}
    \end{minipage}
\end{table}

\begin{table}[ht]
    \centering
    \begin{minipage}[t]{0.45\textwidth}
    \centering
    \begin{tabular}{cccc}
    \toprule
         \textit{\textbf{Client}} & \textit{\textbf{Vanilla FL}} & \textit{\textbf{Global-DP}} & \textit{\textbf{Metric-privacy}} \\
         \midrule
         \textit{Aggregated} & 1.032 & 1.846 & 1.163 \\ 
         \textit{Target}  & 1.182 & 2.681 & 1.548 \\ 
         \midrule
         \textit{Difference ($\%$)} & 12.719 & 31.173 & 24.820 \\ 
    \bottomrule
    \end{tabular}
    \caption{Loss obtained for the aggregated test set and for the shadow dataset of client 3 with \textit{FedProx} strategy.}
    \label{tab:ciattacks_example_fedprox}
    \end{minipage}
    \hspace{0.05\textwidth} 
    \begin{minipage}[t]{0.45\textwidth}
    \centering
    \begin{tabular}{cccc}
    \toprule
         \textit{\textbf{Client}} & \textit{\textbf{Vanilla FL}} & \textit{\textbf{Global-DP}} & \textit{\textbf{Metric-privacy}} \\
         \midrule
         \textit{Aggregated} & 0.795 & 0.949 & 0.815 \\ 
         \textit{Target}  & 0.895 & 1.170 & 0.957 \\ 
         \midrule
         \textit{Difference ($\%$)} & 11.150 & 18.861 & 14.888 \\ 
    \bottomrule
    \end{tabular}
    \caption{Loss obtained for the aggregated test set and for the shadow dataset of client 3 with \textit{FedYogi} strategy.}
    \label{tab:ciattacks_example_fedyogi}
    \end{minipage}
\end{table}

\section{Code reproducibility}\label{sec:software}

For carrying out the experiments conducted in this work, the following Python libraries and versions with their corresponding dependencies have been used (only the main ones are detailed): flower (v1.13.0), TensorFlow (v2.14.0), keras (v2.14.0),  scikit-learn (v1.5.2), scipy (v1.14.1), tqdm (v4.66.5), opencv-python (v4.10.0.84), numpy (v1.26.0), pandas (v2.2.3), pillow (v11.0.0) and matplotlib (v3.9.2). 

\section{Including server side metric-Differential Privacy in Flower}\label{sec:flwr}

In this section we explain further details concerning the code implemented and the software used for carrying out the Federated Learning and for including both global-DP and metric-privacy from the server side. Note that for performing the FL scenario we have used the Flower Python library \cite{beutel2020flower}. This library allows to include DP in a FL workflow (using \cite{geyer2017differentially} and \cite{mcmahan2017learning} as references) using different wrappers, both from the client and server side. From the server side, it implements a wrapper that can be applied to each aggregation strategy for global-DP with server-side fixed and adaptive clipping. In the study conducted in this work we have applied fixed clipping. The Gaussian mechanism is used in the class \textit{DifferentialPrivacyServerSideFixedClipping}, which receives as input the noise multiplier, the clipping norm and the number of clients samples, as explained in Section~\ref{sec:methodology}. Usually a value greater than 1 is recommended for the noise multiplier, but in our use case we analyzed different values in order to select one that allow the model to converge. In particular, with the value 0.01 we achieved convergence and at the same time it allows us to protect against the client inference attacks that we are trying to prevent (as shown in Section~\ref{sec:results_ciattacks}). 

To add the notion of metric DP, we have modified the code of the class \textit{DifferentialPrivacyServerSideFixedClipping} if Flower in such a way that a new function (named \textit{distance\_metric}) is added for dynamically calculate in each round the maximum distance between the different pairs of models as explained in Section~\ref{sec:methodology}. This function receives the results containing a list with the tuple \textit{(ClientProxy, FitRes)}.

Specifically, within the function \textit{aggregate\_fit}, a call to \textit{distance\_metric}) has been included to calculate the distance once the clipping is performed, and then this distance is entered by dividing the noise multiplier in the call to the \textit{add\_gaussian\_noise\_to\_params} and \textit{compute\_stdv} functions for both adding the noise and calculating the standard deviation of the mechanism applied.

These modifications have been done in the file \textit{dp\_fixed\_clipping.py} that is located under the following structure:

\newpage
\dirtree{%
.1 flower.
.2 $\hdots$.
.2 datasets.
.2 dev.
.2 doc.
.2 e2e.
.2 examples.
.2 glossary.
.2 \textcolor{red}{src}.
.3 cc/flwr.
.3 docker.
.3 kotlin.
.3 proto/flwr/proto.
.3 \textcolor{red}{py}.
.4 \textcolor{red}{flwr}.
.5 cli.
.5 client.
.5 common.
.5 proto.
.5 \textcolor{red}{server}.
.6 compat.
.6 driver.
.6 serverapp.
.6 \textcolor{red}{strategy}.
.7 $\hdots$.
.7 \textcolor{red}{dp\_fixed\_clipping.py}.
.7 $\hdots$.
.6 $\hdots$.
.5 $\hdots$.
.4 flwr\_tool.
.3 swift/flwr.
.2 $\hdots$.
}

\end{document}